\documentclass[10pt,twocolumn,letterpaper]{article}

\usepackage{iccv}
\usepackage{times}
\usepackage{epsfig}
\usepackage{graphicx}
\usepackage{amsmath}
\usepackage{amssymb}
\usepackage{bm}
\usepackage{soul}
\usepackage{multirow}
\usepackage{booktabs}
\usepackage{bbm}
\usepackage[dvipsnames]{xcolor}
\usepackage{caption}
\usepackage{adjustbox}
\usepackage{tikz}
\usepackage{makecell}
\usepackage{tabu}
\usepackage{enumitem}
\usepackage[accsupp]{axessibility}

\makeatletter
\NewDocumentCommand{\MakeTitleInner}{ +m +m +m }{
    \newpage%
    \null%
    \vskip 2em%
    \begin{center}%
        \let \footnote \thanks
        {\LARGE #1 \par}
        \vskip 1.5em%
        {%
            \large
            \lineskip .5em%
            \begin{tabular}[t]{c}%
                #2
            \end{tabular}\par%
        }%
        \vskip 1em%
        {\large #3}
    \end{center}%
    \par
    \vskip 1.5em%
}
\NewDocumentCommand{\MakeTitle}{ +m +m +m }{%
    \begingroup
        \renewcommand\thefootnote{\@fnsymbol\c@footnote}%
        \def\@makefnmark{\rlap{\@textsuperscript{\normalfont\@thefnmark}}}%
        \long\def\@makefntext##1{\parindent 1em\noindent
            \hb@xt@1.8em{%
                \hss\@textsuperscript{\normalfont\@thefnmark}%
            }##1%
        }%
        \if@twocolumn
            \ifnum \col@number=\@ne
                \MakeTitleInner{#1}{#2}{#3}
            \else
                \twocolumn[\MakeTitleInner{#1}{#2}{#3}]%
            \fi
        \else
            \newpage
            \global\@topnum\z@   
            \MakeTitleInner{#1}{#2}{#3}
        \fi
        \thispagestyle{plain}\@thanks
    \endgroup
    \setcounter{footnote}{0}%
}
\makeatother

\usepackage{pifont}
\usepackage{alphalph}

\usepackage[pagebackref=true,breaklinks=true,letterpaper=true,colorlinks,bookmarks=false]{hyperref}

\iccvfinalcopy 

\def\iccvPaperID{11008} 

\ificcvfinal\fi

\newcommand{\ours}{\textsc{CrossLoc3D}}
\newcommand{\oursdata}{\textsc{CS-Campus3D}}

\begin{document}

\title{ 
\textsc{CrossLoc3D}:
Aerial-Ground Cross-Source 3D Place Recognition 
}


\author{
Tianrui Guan$^{1}$
~~~ Aswath Muthuselvam$^{1}$ 
~~~ Montana Hoover$^{1}$
~~~ Xijun Wang $^{1}$ \\
~~~ Jing Liang $^{1}$
~~~ Adarsh Jagan Sathyamoorthy $^{1}$ 
~~~ Damon Conover $^{2}$
~~~ Dinesh Manocha$^{1}$\\ \\
$^1$University of Maryland, College Park ~~~~~
$^2$DEVCOM Army Research Laboratory\\
}


\maketitle

\ificcvfinal\thispagestyle{empty}\fi

\begin{abstract}
We present \ours{}, a novel 3D place recognition method that solves a large-scale point matching problem in a cross-source setting. Cross-source point cloud data corresponds to point sets captured by depth sensors with different accuracies or from different distances and perspectives. We address the challenges in terms of developing 3D place recognition methods that account for the representation gap between points captured by different sources. Our method handles cross-source data by utilizing multi-grained features and selecting convolution kernel sizes that correspond to most prominent features. Inspired by the diffusion models, our method uses a novel iterative refinement process that gradually shifts the embedding spaces from different sources to a single canonical space for better metric learning. In addition, we present \oursdata{}, the first 3D aerial-ground cross-source dataset consisting of point cloud data from both aerial and ground LiDAR scans. The point clouds in \oursdata{} have representation gaps and other features like different views, point densities, and noise patterns. We show that our \ours{} algorithm can achieve an improvement of 4.74\% - 15.37\% in terms of the top 1 average recall on our \oursdata{} benchmark and achieves performance comparable to state-of-the-art 3D place recognition method on the Oxford RobotCar. The code and \oursdata{} benchmark will be available at \href{https://github.com/rayguan97/crossloc3d}{github.com/rayguan97/crossloc3d} .
\end{abstract}



\vspace{-10pt}
\section{Introduction}

Place recognition is an important problem in computer vision, with a wide range of applications including SLAM~\cite{orbslam}, autonomous driving~\cite{autodriving1}, and robot navigation~\cite{roboticNav}, especially in a GPS-denied region. 
Given a point cloud query, the place recognition task will predict an embedding such that the query will be close to the most structurally similar point clouds in the embedding space.
The goal of place recognition is to compress information from a database with a location tag and find the closest data point from the database given a query, which is essentially an information retrieval task on a global scale.

\begin{figure}[t]
    \centering
    \includegraphics[width=\columnwidth]{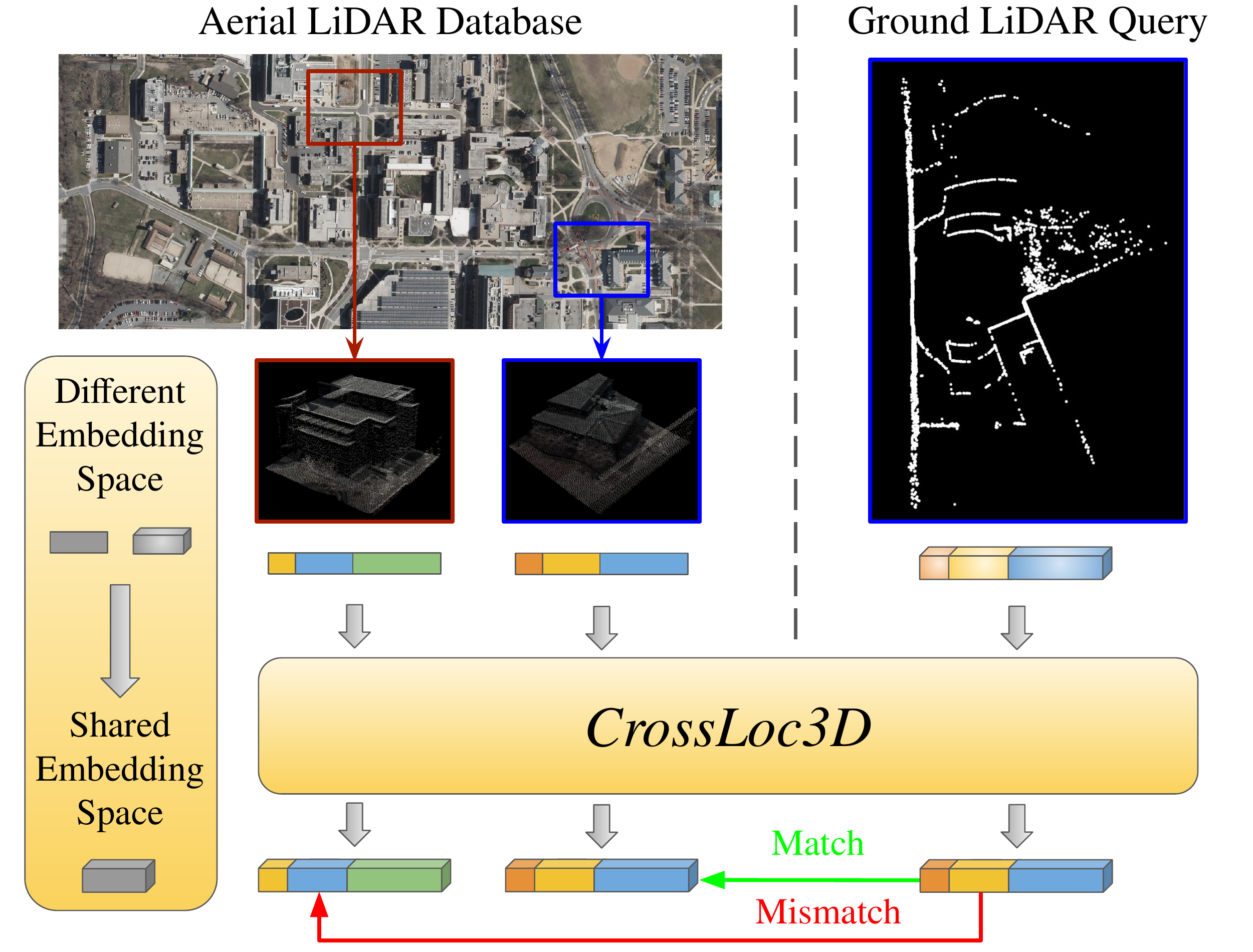}
    \vspace{-3mm}
    \caption{\textbf{\ours{}:} Our method processes cross-source point clouds into a better, shared embedding, which achieves a better retrieval outcome in a cross-source setting.}
    \label{fig:cover_net}
    \vspace{-7mm}
\end{figure}

\begin{figure*}[t]
    \centering
    \includegraphics[width=\textwidth]{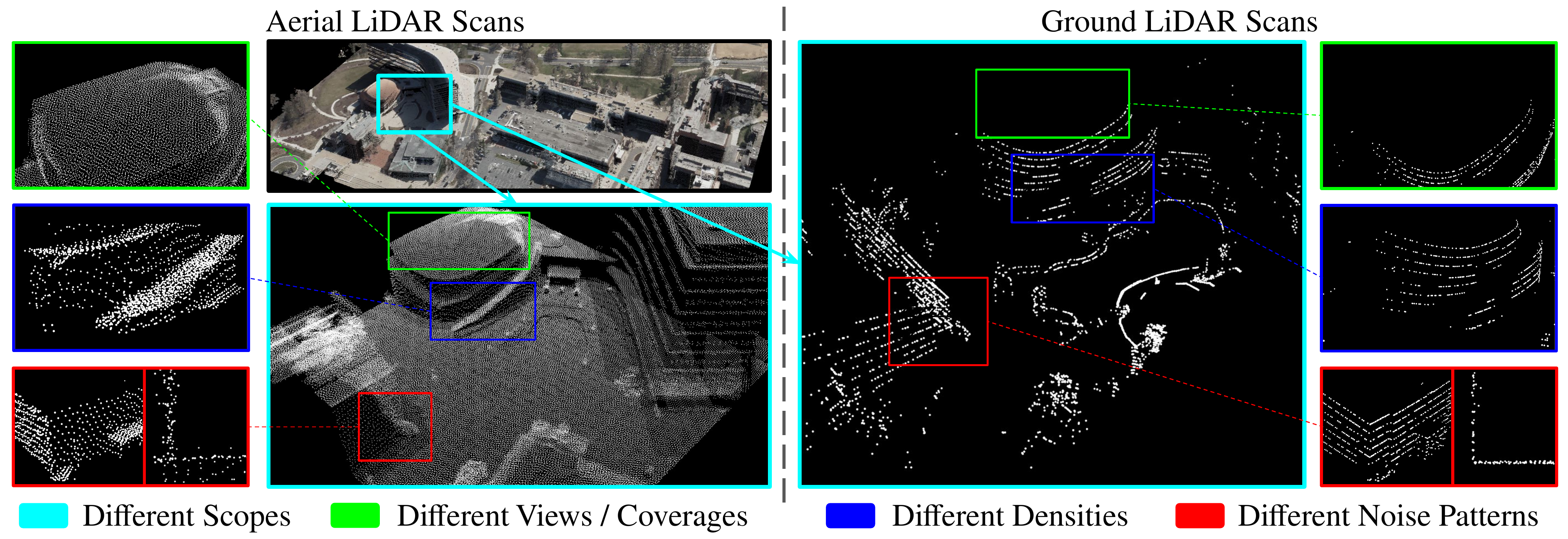}
    \vspace{-6mm}
    \caption{\textbf{Representation gap between aerial and ground sources:} We use the bounding box with the same color to focus on the same region and highlight the differences between aerial (left) and ground (right) LiDAR scans. \textbf{Scopes (\textcolor{cyan}{cyan}):} The aerial scans cover a large region, while ground scans cover only a local area. \textbf{Coverages (\textcolor{green}{green}):} The aerial scans cover the top of the buildings, while ground scans cover more details on the ground.
    \textbf{Densities (\textcolor{blue}{blue}):} The distribution and density of the points are different because of various scan patterns, effective ranges, and fidelity of LiDARs.
    \textbf{Noise Patterns (\textcolor{red}{red}):} The aerial scans have larger noises, as we can see from a bird-eye view and top-down view of a corner of the building. 
    }
    \label{fig:cover}
    \vspace{-18pt}
\end{figure*}



In this paper, we deal with the problem of 3D cross-source place recognition, as illustrated in Fig.~\ref{fig:cover_net}.
The goal of 3D cross-source place recognition in the context of aerial views and ground views is extremely challenging and not well-studied in the field~\cite{crosssource_survey}. There are two aspects of cross-source data that result in additional challenges: cross-view and data consistency. First, the perspective differences between aerial and ground datasets would cause partial overlap of the scans, leading to a lack of point correspondences. Second, there may be no data consistency between sources, which will cause representation gaps. The challenge of using multiple sources of data is primarily related to the different nature and fidelity of sensors.
It is harder to match the 3D point clouds from different sources~\cite{crosssource_survey} than it is to perform similar matches between the 2D images captured from the same locations, or even point clouds captured only by ground or aerial sensors. 
When different kinds of LiDAR sensors or the same kind of LiDAR at different distances, capture scans at the same location, the data is significantly different. 3D cross-source data is point cloud data captured by different depth sensors and differ significantly in terms of scope, coverage, point density, and noise distribution pattern, as shown in Fig.~\ref{fig:cover}. 
We need a method that can better close the representation gap between the two sources by converting features in different embedding spaces to a common embedding space.

While the problems of localization and point-set registrations are well-studied, there is relatively less work on cross-source localization.  Ge et al.~\cite{local_to_global_semantic} propose a  localization method using point cloud data from both the air and the ground sensors. However, this method relies on semantic information for accurate 2D template matching; in addition, their data is private. 
To the best of our knowledge, there are no existing works that only utilize raw point information obtained from both ground and aerial LiDAR sensors and there are no known open datasets for such applications.



\noindent{\bf Main Results:} In this paper, we propose a novel 3D place recognition method that works well on both single-source and cross-source point cloud data.  We account for the representation gap by using multi-grained features and selecting appropriate convolution kernel sizes.  Our approach is  inspired by the diffusion model~\cite{ddpm, ddim} and cold diffusion~\cite{cold_diff} and we propose a novel iterative process to refine multi-grained features from coarse to fine. We also propose a novel benchmark dataset that consists of point cloud data from both ground and aerial views. The novel aspects of our approach include:

\begin{enumerate}
    
    \item We propose \ours{}, a novel place recognition method that utilizes multi-grained voxelization and multi-scale sparse convolution with a feature selection module to actively choose useful features and close the representation gap between different data sources.
    
    \item We propose an iterative refinement process that shifts the feature distributions of various input sources toward a canonical latent space. We show that starting the refinement process from the coarsest features, which are most similar across different sources, toward the finer features, leads to better recall compared to doing it in the reverse order or simply concatenating features from different resolutions.
    
    \item We present the first public 3D Aerial-Ground Cross-Source benchmark in a campus environment, \oursdata{}, which consists of both aerial and ground LiDAR scans. We collect ground LiDAR data on mobile robots and process the aerial data from the state government
    into a suitable format to cross-reference the ground data. The dataset and code will be publicly released and made available for benchmark purposes.
    \end{enumerate}
We  have evaluated \ours{} and other state-of-the-art 3D place recognition methods on the CS-Campus3D dataset and observe an improvement of 4.74\% - 15.37\% in terms of the top 1 average recall. Additionally, we observe that \ours{} achieves close to 99\% in terms of top 1\% average recall on Oxford RobotCar~\cite{oxford} and performs comparably, within a margin of 0.31\%, to the SOTA methods on the traditional single-source 3D place recognition task.


\section{Related Work}

\subsection{2D and 3D Place Recognition}

Place recognition is finding a match for a query position among a database of locations based on its traits and features. Image-based place recognition~\cite{vpr_survey} has been well-studied using Vector of Locally Aggregated Descriptor (VLAD) approaches~\cite{vlad1, vlad2, netvlad} and has been extended into 3D domains. As the first point-based method for large-scale place recognition, PointNetVLAD~\cite{pointnetvlad} uses PointNet~\cite{pointnet} as a feature extractor for the NetVLAD~\cite{netvlad} and has been evaluated on the Oxford RobotCar~\cite{oxford} dataset. Since the creation of this benchmark, many other methods have been developed with improved performance~\cite{dh3d, minkloc3d, soenet, svtnet, pptnet}.

\subsection{Multi-View Localization}

Multi-view localization utilizes different perspectives as inputs for localization and has been widely studied, especially in the context of image-based geo-localization~\cite{crossview1, crossview2, crossview3, crossview4, xview, vis-cross-view}. Gawel et al.~\cite{xview} use semantically segmented images from both the air and the ground to build a semantic graph for localization on a global map. 
Xia et al.~\cite{vis-cross-view} approach such cross-view localization tasks from the probabilistic perspective on the satellite image instead of as an image retrieval objective. 
\cite{aerial_img_ground_lidar} proposes a method to combine local LiDAR sensors with an occupancy map obtained from overhead satellite imagery for place recognition and pose estimation. 
In addition, Ge et al.~\cite{local_to_global_semantic} convert both aerial and ground LiDAR into 2D maps and utilize semantic and geometric information for 2D template matching.

\subsection{3D Cross-Source Matching}

3D cross-source matching has been studied under the context of point-level registration. Huang et al.~\cite{crosssource_survey} present a 3D cross-source registration dataset consisting of indoor point clouds captured from either Kinect or LiDAR or calculated from 3D reconstruction algorithms from 2D images~\cite{vsfm} as different sources. This benchmark poses many challenges due to noise, partial overlap, density difference, and scale difference. To deal with the lack of correspondences caused by noise and density differences, Huang et al.~\cite{fmr} propose a semi-supervised approach to improve the robustness of the registration.
In the presence of partial overlap and scale difference of the two matching point clouds, Huang et al.~\cite{coarse-to-fine} propose a coarse-to-fine algorithm to gradually finalize the candidate of the match.
In our context, the challenges are similar in terms of the points representation gap, but our problem introduces more difficulties due to the discrepancy caused by perspective differences in the data captured from aerial and ground viewpoints. In addition, the existing cross-source benchmark~\cite{crosssource_survey} focuses on point-level registration of indoor objects, while our benchmark focuses on outdoor large-scale place recognition.

\section{Problem Definition}



Let query $\mathcal{Q}_g$ be a set of 3D points in a single frame captured by a LiDAR sensor $L_g$ from the ground represented in its relative coordinate. The range radius of $L_g$ is $r$.

Let $\mathcal{M}_a$ be a large-scale 3D point cloud map consisting of multiple frames captured by a LiDAR sensor $L_a$ from the aerial viewpoint. The global point cloud map $\mathcal{M}_a$ is divided into a collection of $M$ overlapping submaps $\mathcal{D}_a=\{ m_1, m_2, ... m_M \}$, where the center of the submaps is uniformly spaced at a distance of $d$ and with an area of coverage (AOC) of $A$. The parameters $d$ and $A$ are chosen so that $|m_i| \ll |\mathcal{M}_a|$ and $A \approx \pi r^2$.

\noindent\textbf{Definition.} \textit{Given a query $\mathcal{Q}_g$ and a database $\mathcal{D}_a$, the goal is to retrieve the submap $m_i\in \mathcal{D}_a$ whose coverage includes the location $L$ at which $\mathcal{Q}_g$ is captured in map $\mathcal{M}_a$. 
}

\noindent\textbf{Approach.} Our network takes a set of points $\mathcal{Q}_g$ as input and learns a function $f(.)$ that maps $\mathcal{Q}_g$ to a feature vector $v_g$. For $m_i, m_j\in \mathcal{D}_a$, we expect $f(.)$ to satisfy $d(f(\mathcal{Q}_g), f(m_i)) < d(f(\mathcal{Q}_g), f(m_j))$ if $\mathcal{Q}_g$ is structurally similar to $m_i$ but dissimilar to $m_j$. The 3D place recognition task can be defined as finding the submap $m_*\in\mathcal{D}_a$ such that $ d(f(\mathcal{Q}_g), f(m_*)) < d(f(\mathcal{Q}_g), f(m_i)), \forall i \neq *$. 

\noindent\textbf{Unique Setting.} Compared to previous 3D place recognition tasks in which all data are collected from the same platform, 
our data are mixed with two different sources from both aerial scans and ground scans, which leads to cross-localization problems. This setting poses a unique challenge due to the differences in cross-source data as indicated in Fig.~\ref{fig:cover}.
Under this setting, most existing 3D place recognition methods ~\cite{pointnetvlad, lpdnet, minkloc3dsi, transloc3d} lead to poor performance and low recall, as shown in Table.~\ref{tab:comparisons_ours}.

\begin{figure*}[t]
    \centering
    \includegraphics[width=\textwidth]{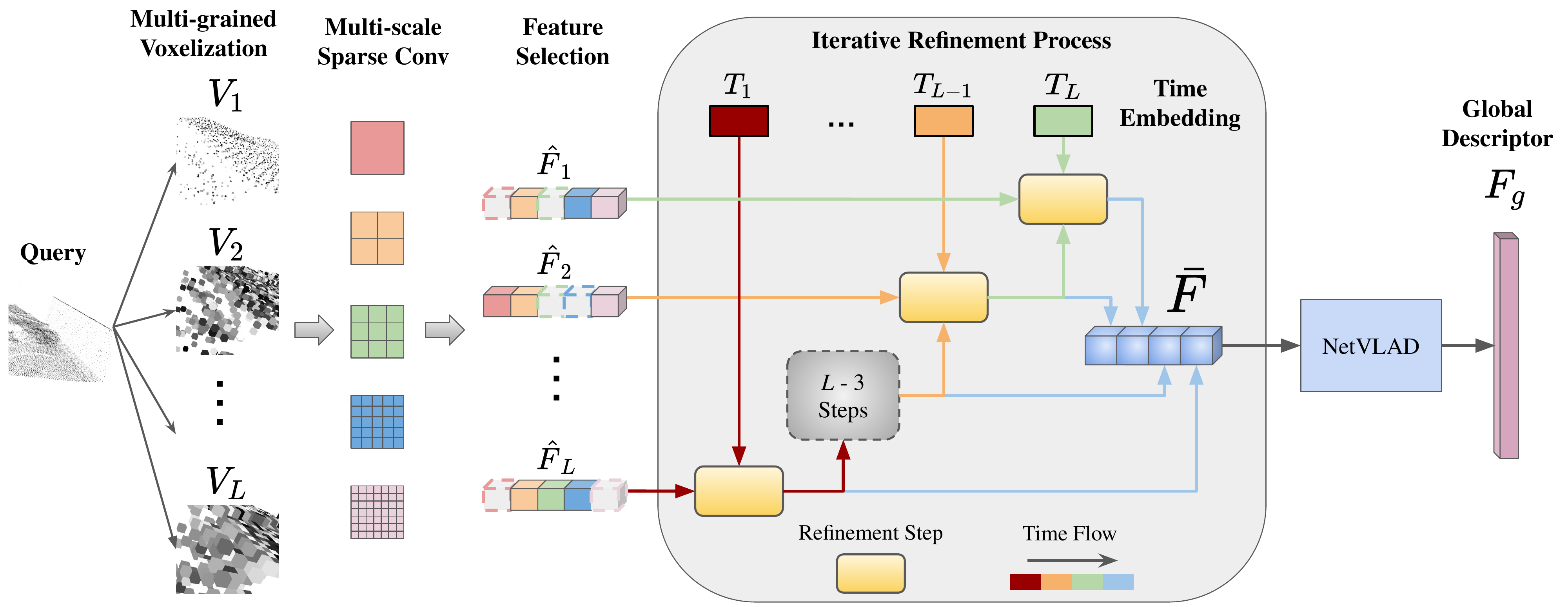}
    \caption{\textbf{Architecture of our network \ours{}:} Our network consists of three parts: multi-grained feature selection, iterative refinement, and NetVLAD. First, we run the points through multi-grained voxelizations and several streams of sparse convolution operations with different kernel sizes. We use feature selection to choose the best field of view and keep the corresponding features $\hat{F}_i$ for each voxelization $V_i$. Second, we perform an iterative refinement process to obtain a set of local features $\bar{F}$. Finally, the NetVLAD~\cite{netvlad} module will produce the global descriptor $F_g$ by aggregating local features $\bar{F}$.
    }
    \label{fig:network}
    \vspace{-18pt}
\end{figure*}

\section{Method}

\subsection{Overview}
Our network, \ours{}, takes a point cloud as input and outputs a global descriptor such that two sets of structurally similar points have closer global descriptors.
Our proposed architecture consists of three stages: 1) multi-grained feature extraction with feature selection, 2) an iterative refinement model accounting for distribution shift between different sources, and 3) a NetVLAD~\cite{netvlad}, as shown in Fig.~\ref{fig:network}. 
The network needs to approximate a function with inputs from both ground and aerial sources, which do not have any exact point correspondence or similar noise distributions.
During training, we use lazy triplet margin loss~\cite{pointnetvlad}, which is commonly used for place recognition.

\subsection{Multi-grained Features with Selection}

We process the point cloud patches into a uniform size and extract their multi-scale features at different voxelization sizes. The purpose of the module is to accommodate data from different sources by obtaining features in different granularities and selecting their respective dominant features.
Let $p\in \{ \{ \mathcal{Q}_g \}, \mathcal{D}_a \} \subseteq \mathbb{R}^{n \times 3}$ be a patch from either ground or aerial sources and down-sampled to size $n$. 

\noindent\textbf{Multi-grained Features:} First, we perform $L$ voxelization on $p$ with different resolutions $\{ r_1, r_2, ..., r_L |\ \forall\ 0 \leq i < j < L,  r_i < r_j  \}$ and obtain a list of voxel sets  $V_1, V_2, ..., V_L$. 
A series of sparse convolution operations, batch normalization, and activation is performed on each voxel set $V_i$. After that, we use sparse convolution with 5 different field of viwes $1, 3, 5, 7, 9$ to obtain a multi-scale feature $F_i \in \mathbb{R}^{5 \times N_i \times D_v}$ from $V_i$, where $N_i$ is the number of voxels in $V_i$ and $D_v$ is the feature dimension. 

\noindent\textbf{Feature Selection:} Given the multi-scale feature $F_i$, where $i=1, 2, ..., L$ for each voxel grain, we may have some redundant information from different quantization sizes and scopes of the kernels. In addition, when the input points are from one source, a specific quantization size coupled with a specific field of view could work better when the input is from another source. As a result, we add another feature selection module that can choose the best field of view for each feature $F_i$ based on a confidence score. We can obtain the feature after selection:
\vspace{-3mm}
\begin{equation}
\hat{F}_i = F_i[\ \textit{topk}(\ \textit{softmax}(\ g(F_i),\  dim = 0),\  k = k_s)\ ] ,
\vspace{-3mm}
\end{equation}
where $g(.)$
is a learnable function by a few layers of convolutions, $k_s$ is the number of features to select ($0 < k_s \leq 5$), and $\hat{F}_i \in \mathbb{R}^{k_s \times N_i \times D_r}$ for some feature size $D_r$. 

Using a large voxel size can lose information, but it provides some flexibility in matching when one source has a different noise pattern or density from another source. With feature selection, we want to keep the best information obtained from multi-grained quantizations and multi-scale kernels.

Note that during inference, the database can be computed ahead of time, and only the query point needs to be passed through the network. 
The increase in run-time can be negligible given the amount of improvement from the proposed \ours{}, as shown in Table.~\ref{tab:complexity}.

\subsection{Iterative Refinement}
\label{sec:iterrefine}
Inspired by diffusion model~\cite{ddpm, ddim, cold_diff}, we hope to close the representation gap between different sources, by gradually learning a distribution shift towards a canonical space at each refinement step in the training process. 
Let $\hat{F} = \textit{permute}([ \hat{F}_1, \hat{F}_2, ..., \hat{F}_{L} ]))$, where $ \hat{F}[i]\in \mathbb{R}^{k_s \times \hat{N}_i \times D_r}$ and $\textit{permute}(.)$ gives control over the order of refinement process. Note that only when the original order is maintained, $\ N_i = \hat{N_i}$ for all $i$. For simplicity of the notations, we assume the ascending order of $\hat{F}$ and $\ N_i = \hat{N_i}$.

Let $r_i(.) : A \to A$ be a refinement function at step $i$, where $L$ is the total number of steps for the refinement and $A \subseteq \mathbb{R}^{k_s \times * \times D_r}$. Let $\tilde{F}_1 = r(\hat{F}[1]) $ be the first step of the refinement, then for $1 < i \leq L$, $\tilde{F}_i$ is defined as follows:
\vspace{-2mm}
\begin{equation}
\tilde{F}_i = r(\textit{concate}(\bigr[\tilde{F}_{i-1}, \hat{F}[i]\bigr],\ dim=1)) ,
\vspace{-2mm}
\end{equation}
where $\tilde{F}_i \in \mathbb{R}^{k_s \times \sum_{j=1}^{i}{N_j} \times D_r}$.

The final output of the refinement at step $i$ is defined as:
\vspace{-2mm}
\begin{equation}
\bar{F}_i = \textit{concate}(\bigr[ \tilde{F}_i, \hat{F}[i+1], ..., \hat{F}[L]  \bigr],\  dim=1)
\vspace{-2mm}
\end{equation}

After $L$ steps of iterative refinement, we can obtain $\bar{F}_i \in \mathbb{R}^{k_s \times N \times D_r}$, where $N = \sum_{j=1}^{L}{N_i}$. We concatenate all $\bar{F}_i$ along the feature dimension and reshape it to get $\bar{F} \in \mathbb{R}^{N \times D_g}$, where $D_g = k_s \times D_r \times L$.

\noindent\textbf{Refinement Function:} $r(.)$ function is implemented with a series of External Attention (EA)~\cite{eanet} blocks, where each block is considered a substep of the refinement. Instead of using features to generate query, key, and value tensors, the EA block only uses a single stream for query input and uses internal memory units to generate attention maps, which leads to strong performance with only linear time complexity. Please refer to EANet~\cite{eanet} for more details.

\noindent\textbf{Time Embedding:} At step $i$, there are multiple substeps within $r_i(.)$. For each substep $j$, we add a learnable sinusoidal positional embedding to the iteration. Let $t = iL + j$, then the positional embedding is defined as follows:
\vspace{-3mm}
\begin{equation}
    PE(t) = \textit{concate}(\bigr[ t, sin(2\pi w t), cos(2\pi w t) \big])\ ,
\vspace{-3mm}
\end{equation}
where $w$ is a learnable parameter. $PE(t)$ is directly added to the input features after a few linear and activation functions.

The diffusion models~\cite{ddpm, ddim} use a deep neural network to learn a small distribution shift between two consecutive steps. Although the fundamental theory of the diffusion process is based on the assumption of Gaussian noise in the forward process, Bansal et al.~\cite{cold_diff} show that the process can be learned even when the degradation is completely deterministic. We want to simulate a deterministic function that shifts different feature distributions caused by different sources to a common canonical space.
Ideally, at the later stage of the refinement, the features from different sources should be in a better embedding space for similarity measurement. 
We provide more analysis in Sec.~\ref{sec:ablation}.

\subsection{Vector of Locally Aggregated Descriptors}

NetVLAD~\cite{netvlad} (Network Vector of Locally Aggregated Descriptors) is a deep neural network architecture that is commonly used in image-based or point-based place recognition tasks. 
Due to its permutation invariant property, the NetVLAD layer takes in a set of local feature descriptors and outputs a fixed-length global descriptor that encodes information about the entire input. 
It learns $K$ cluster centers and uses the sum of residuals as global feature descriptors based on cluster assignment.

Let $\bar{F} = \{ x_1,x_2,\dots,x_N \} \in \mathbb{R}^{N \times D_g}$ be a set of local feature descriptors from the iterative refinement step. Let $C=\{c_1,c_2,\dots,c_K\}\in\mathbb{R}^{K \times D_g}$ be a set of cluster centroids of the same dimension $D_g$. 
The global feature $F_g\in\mathbb{R}^{D_g \times K}$ is computed as follows:
\vspace{-2mm}
\begin{equation}
F_g(j, k) = \sum_{i=1}^{N} a_k(x_i) (x_i(j) - c_k(j)) \ ,
\vspace{-3mm}
\end{equation}
where $a_k(x_i)_ = \frac{e^{-\alpha||x_i-c_k||^2}}{\sum_{k'} e^{-\alpha||x_i-c_{k'}||^2}}$ is the soft assignment function and $\alpha$ is a temperature parameter that controls the softness of the assignment. Note that the NetVLAD layer can be trained end-to-end with similarity loss between the global descriptors of different inputs.

\subsection{Triplet Margin Loss for Metric Learning}

We train the embedding function $f(.)$ with a lazy triplet margin loss~\cite{pointnetvlad}. Let $\mathcal{T} = \{p_A, \{ p_{pos} \}, \{ p_{neg} \} \}$ be a training triplet, where $p_A$, $\{ p_{pos} \}$ and $\{ p_{neg} \}$ denote the anchor point cloud, a set of positive point clouds that are structurally similar to $p_a$, and a set of negative point clouds that are dissimilar to $p_a$, respectively. For more efficient training, we use a hard example mining technique~\cite{mining} to get hard positives and negatives in the triplet. 
The lazy triplet loss is formulated as follows: 
\vspace{-3mm}
\begin{equation}
\mathcal{L}_{\text{lazy}}(\mathcal{T},\alpha) = \max(0, \delta_{pos} - \delta_{neg} + \alpha),
\vspace{-3mm}
\end{equation}
where $\alpha$ is a constant margin parameter, $\delta_{pos}=D(f(p_A), f(p_{pos}))$ and  $\delta_{neg}=D(f(p_A), f(p_{neg}))$ for some distance function $D$.

Based on the lazy triplet margin loss, the similarity of the point cloud is based on the distance of the output embedding defined by distance function $D$. During inference, we calculate the distance between the database embedding and query embedding to retrieve the predicted neighbors based on the ranking of the distance.

\section{CS-Campus3D Dataset}
\label{sec:dataset}
In this section, we present a novel \oursdata{} benchmark for cross-source 3D place recognition task.
The CS-Campus3D benchmark consists of both aerial and ground LiDAR patches, which are tagged with UTM coordinates at their respective centroids. The coordinates of the patches are shifted and scaled to $[-1, 1]$, and the points in each patch are randomly down-sampled to size $n=4096$. In Table.~\ref{tab:dataset}, we compare our dataset with a traditional 3D place recognition benchmark dataset, Oxford RobotCar.

\begin{table}[t]
\centering
\vspace{-8 pt}
\resizebox{\columnwidth}{!}{%
\begin{tabular}{|c|c|c|}
\hline
    \textbf{Dataset} & \textbf{Oxford RobotCar~\cite{oxford}}
    & \textbf{\rotatebox[origin=c]{0}{\makecell{\oursdata{} \\ (Ours)}}}\\
\hline
     Scenarios & \rotatebox[origin=c]{0}{\makecell{Urban,\\ Suburban}}  & Campus  \\
\hline
     Platform & \rotatebox[origin=c]{0}{\makecell{Nissan LEAF (Car) \\ }}  & \rotatebox[origin=c]{0}{\makecell{Spot, Husky \\(Mobile robots)\\ }}  \\
\hline
     Road & Car Lane  & \rotatebox[origin=c]{0}{\makecell{Car Lane, Alley,\\ Side Walk, etc.\\ }}  \\
\hline
     \rotatebox[origin=c]{0}{\makecell{Perception \\ Sensors}} & RGB Camera, LiDAR  & LiDAR  \\
\hline
     \rotatebox[origin=c]{0}{\makecell{Geographical \\ Coverage}} & \rotatebox[origin=c]{0}{\makecell{$\sim 10\ km$ \\ (Driving distance)}}    & \rotatebox[origin=c]{0}{\makecell{$1628\times 3377\ m$ \\ (Area)}}   \\
\hline
     \rotatebox[origin=c]{0}{\makecell{Num. of Ground \\ Submaps \\ (training/testing)}} & $21711/3030$  & $6167/1538$  \\
\hline
     \rotatebox[origin=c]{0}{\makecell{Num. of Aerial \\ Submaps \\ (training/testing)}} & $N/A$  &  $27520/0$ \\
\hline
     Patch Size &  $\sim10-20\ m$  &  $\sim100\ m$ \\
\hline
     \rotatebox[origin=c]{0}{\makecell{Positive Pair \\ Distance}}   & $25\ m$  & $100\ m$  \\
\hline

\end{tabular}
}
\caption{\textbf{Our proposed benchmark \oursdata{}:} We give a side-by-side comparison between the most popular 3D place recognition benchmark Oxford RobotCar and our proposed dataset \oursdata{}. We highlight that our benchmark covers more areas due to the flexibility of mobile robots and contains LiDAR data from both ground and aerial sources. We use those two datasets for evaluation. }
\label{tab:dataset}
\vspace{-15pt}
\end{table}

\noindent\textbf{Aerial LiDAR Scan:}
The aerial data is obtained from the state government and is publicly available~\cite{mdmap}. The dataset is collected by an airborne LiDAR sensor mounted on an airplane. The points are in UTM coordinates and point spacing is roughly $0.35\ m$. The vertical and horizontal accuracies are $0.3\ m$ and $0.1\ m$, respectively. 

The state-wide aerial LiDAR scan is trimmed down to a $1628\times 3377\ m$ region that covers the same area as the ground scan does, and ground surfaces are removed. To construct the database, we uniformly sample $100\times 100\ m$ patches several times every $19\ m$, moving north and east, and obtained 27520 patches from aerial data. The patch size is chosen according to the effective range of our LiDAR. 

\noindent\textbf{Ground LiDAR Scan:} The ground LiDAR scan is collected on a Boston Dynamic Spot and a Clearpath Husky equipped with a Velodyne VLP 16 LiDAR with an effective range of around $100\ m$ and an accuracy of $\pm 3\ cm$.
We tele-operate the robot in the region with corresponding aerial coverage and save the points from each frame. We run 27 different routes with partially overlapping and non-overlapping locations and obtain a total of 7705 patches from the ground data.

\noindent\textbf{Ground Truth Correspondence:} We use a U-blox NEO-M9N GPS module to get the latitude and longitude coordinates of the current LiDAR frame. We convert the geographic coordinates to UTM coordinates to match with the LiDAR data. The accuracy of the GPS is around $1.5\ m$.

\noindent\textbf{Evaluations:} In Oxford RobotCar~\cite{oxford}, the patch size is between $10 - 20\ m$ and the successful retrieval distance is within $25\ m$. Following that ratio, we set a successful retrieval distance to be $100\ m$, which means the query point cloud is successfully localized if the retrieved neighbor is within $100\ m$. We have created two evaluation settings: 
\begin{enumerate}[itemsep=1pt,parsep=1pt, topsep=0pt]
    \item We construct the database set with all aerial LiDAR scans and a subset of ground LiDAR scans for training; we use the rest of the ground data as queries for testing.
    \item For a more challenging setting, we construct the database set with only aerial LiDAR scans and only use ground LiDAR scans as queries. The ground scans will have to find a match only based on aerial scans.
\end{enumerate}
We use the same evaluation metrics as~\cite{pointnetvlad, minkloc3d}, explained in Sec.~\ref{sec:metrics}. The \oursdata{} database, benchmark, and all processing scripts will be publicly released. 


\begin{figure*}[t]
    \centering
    \includegraphics[width=\textwidth]{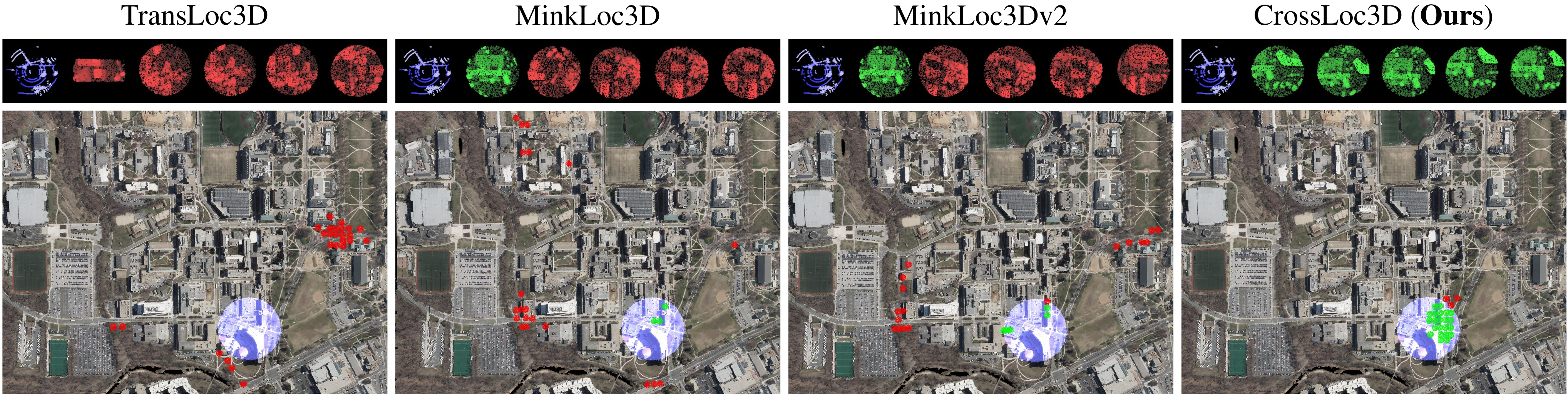}
    \caption{\textbf{Qualitative comparisons between \ours{} and other SOTA methods:} \textbf{Top:} Input ground query (\textcolor{blue}{blue}) and top 5 retrievals (ranked from left to right) from database of aerial data (true neighbor -- \textcolor{green}{green}, false neighbor -- \textcolor{red}{red}). \textbf{Bottom:} Distributions of ground query and top 25 retrievals. 
    }
    \label{fig:inference}
    \vspace{-19pt}
\end{figure*}
\section{Experiments and Evaluations}
\subsection{Dataset and Evaluation Metric}
\label{sec:metrics}

We use the Oxford RobotCar~\cite{oxford} dataset and \oursdata{} (ours) for evaluation. 
Oxford RobotCar is collected on a vehicle with LiDAR sensors while driving on urban and suburban roads in various weather conditions. It has been the benchmark dataset for most of the 3D place recognition methods because the evaluation is justified on this benchmark. Since Oxford RobotCar has a wide coverage of the city as well as repetitive routes under different conditions at different times, the point cloud retrieval task must rely on the similarity between the structures instead of exact point correspondence. The proposed benchmark, \oursdata{}, also satisfies this condition as the patches are disjoint and collected from different sources. For more details about the datasets, please refer to Table.~\ref{tab:dataset} and Sec.~\ref{sec:dataset}.

\noindent\textbf{Evaluation Metrics:} Like most of the place recognition benchmarks~\cite{pointnetvlad, minkloc3d}, we used \textit{AverageRecall@N} and \textit{AverageRecall@1\%} as evaluation metrics. \textit{Recall@N} is defined as the percentage of true neighbors that are correctly retrieved based on the top $N$ predicted neighbors from the network. \textit{Recall@1\%} is equivalent to \textit{Recall@N} when $N$ is equal to $1\%$ of the database size.

\subsection{Implementation Details}

\noindent\textbf{\ours{}: } The input point size $n$ is 4096. We set $L=3$, where the voxelization size is $[0.05, 0.12, 0.4]$. The sparse convolution is implemented based on MinkowskiEngine~\cite{minkowski}. The feature section $k_s$ is 3. The order of the refinement is reversed (coarse $\rightarrow$ fine). During refinement, we set the number of sub-steps to 3 for all $L$ steps. The initial dimension of the time embedding is 8. The feature dimensions of the refinement and NetVLAD are 512 and 256, respectively. We use Euclidean distance for the distance function $D$ during training and inference.

\noindent\textbf{Training Parameters:} Models are trained on a single NVIDIA RTX A5000 for 200 epochs. We use the Adam optimizer with a learning rate of $0.005$ and $(\beta_1, \beta_2) = (0.9, 0.999)$. We use an expansion batch sampler where the initial batch size is 64 and can grow up to 128.

\begin{table}[t]
\centering
\Large
\begin{adjustbox}{max width=\linewidth}
\begin{tabular}{l|c|ccc}
  \toprule[2pt]
  \textbf{Methods} & \textbf{Reference} &
   \textbf{AR@1\%} $\uparrow$ & \textbf{AR@1} $\uparrow$\\
\hline
      PointNetVLAD~\cite{pointnetvlad} & CVPR 2018 &  80.3 & -  \\
      PCAN~\cite{pcan} & CVPR 2019 & 83.8 & - \\

      LPD-Net~\cite{lpdnet}  & ICCV 2019 & 94.9 & 86.3 \\
      DH3D~\cite{dh3d} & ECCV 2020 & 85.3 & 74.2 \\
      PPT-Net~\cite{pptnet}  & ICCV 2021 & 98.1 & 93.5 \\
      SOE-Net~\cite{soenet}  & CVPR 2021 & 96.4 & - \\
     Minkloc3D~\cite{minkloc3d} & WACV 2021 & 97.9 & 93.0 \\    
      Minkloc3Dv2~\cite{minkloc3dv2} & ICPR 2022 & \textbf{98.9} & \textbf{96.3} \\
      Minkloc3D-S~\cite{minkloc3dsi} & RAL 2021 & 93.1 & 82.0 \\    
      SVT-Net~\cite{svtnet}  & AAAI 2022 & 97.8 & 93.7 \\
      TransLoc3D~\cite{transloc3d}  & preprint & 98.5 & \underline{95.0} \\

    \cmidrule{1-4}
      \ours{} \textbf{(Ours)} & - & \underline{98.59} & 94.36 \\




  \bottomrule[2 pt]
\end{tabular}

\end{adjustbox}
\caption{\textbf{State-of-the-art comparisons on Oxford RobotCar:} We compare \ours{} with other state-of-the-art methods on single-source 3D place recognition benchmark. All numbers are obtained from the original paper. We underscore the second best method. 
}
\vspace{-6mm}
\label{tab:comparisons}
\end{table}

\subsection{Comparisons}

\noindent\textbf{SOTA Comparisons:} In Table.~\ref{tab:comparisons} and Table.~\ref{tab:comparisons_ours}, we show the performance of the proposed \ours{} and other SOTA methods. On two evaluation sets of \oursdata{}, \ours{} achieves an improvement of $0.52 - 3.94\%$ and $4.74 - 15.37\%$ in terms of \textit{AR@1\%} and \textit{AR@1}, respectively. We see a trend of improvement in \textit{AR@1\%} when the database only contains aerial data because the total database size decreases. In addition, \ours{} has comparable performance to the state-of-the-art method within a margin of $0.31$\% in terms of $1\%$ average recall. In Fig.~\ref{fig:inference}, we can see that all top 5 retrievals and most of the top 25 retrievals by \ours{} are within the range of true neighbors, while the predictions of other SOTA methods are distributed away from the ground query. 

\begin{table}[t]
\centering
\Large
\begin{adjustbox}{max width=\linewidth}
\begin{tabular}{c|c|ccc}
  \toprule[2pt]
  \textbf{\rotatebox[origin=c]{0}{\makecell{ Evaluation Set: \\ (Database / Query)}}}  & \textbf{Methods} &
   \textbf{AR@1\%} $\uparrow$ & \textbf{AR@1} $\uparrow$\\
\hline
      \multirow{7}{*}{\rotatebox[origin=c]{0}{\makecell{  Aerial + Ground  \\ / \\ Ground-Only  }}} & PointNetVLAD~\cite{pointnetvlad} &  41.46 & 35.57  \\

        & LPD-Net~\cite{lpdnet} & 56.49 & 45.94 \\
      &Minkloc3D~\cite{minkloc3d} & \underline{79.10} & \underline{69.38} \\    
       & Minkloc3Dv2~\cite{minkloc3dv2} & 76.68 & 67.06 \\
       & Minkloc3D-S~\cite{minkloc3dsi} & 59.33 & 44.39 \\    
        & TransLoc3D~\cite{transloc3d} & 69.04 & 58.16 \\
    \cmidrule{2-4}

     &  \ours{} \textbf{(Ours)} & \textbf{83.04} & \textbf{74.12} \\

    \midrule
        \multirow{7}{*}{\rotatebox[origin=c]{0}{\makecell{  Aerial-Only  \\ / \\ Ground-Only }}} & PointNetVLAD~\cite{pointnetvlad} &  43.53 & 19.07  \\

        & LPD-Net~\cite{lpdnet} & 40.70 & 11.99 \\
       &Minkloc3D~\cite{minkloc3d} & \underline{85.22} & \underline{55.36} \\    
       & Minkloc3Dv2~\cite{minkloc3dv2} & 83.48 & 52.46 \\
       & Minkloc3D-S~\cite{minkloc3dsi} & 71.88 & 32.17 \\    
        & TransLoc3D~\cite{transloc3d} & 80.64 & 42.97 \\
    \cmidrule{2-4}

      &  \ours{} \textbf{(Ours)} & \textbf{85.74} & \textbf{70.73} \\

  \bottomrule[2 pt]
\end{tabular}

\end{adjustbox}
\caption{\textbf{State-of-the-art comparisons on the proposed \oursdata{} test set:} We compare \ours{} with other state-of-the-art methods on the cross-source 3D place recognition benchmark in two different settings. We underscore the second best method.
}
\vspace{-7mm}
\label{tab:comparisons_ours}
\end{table}

\begin{figure}[t]
    \centering
    \vspace{-1mm}
    \includegraphics[width=0.95\columnwidth]{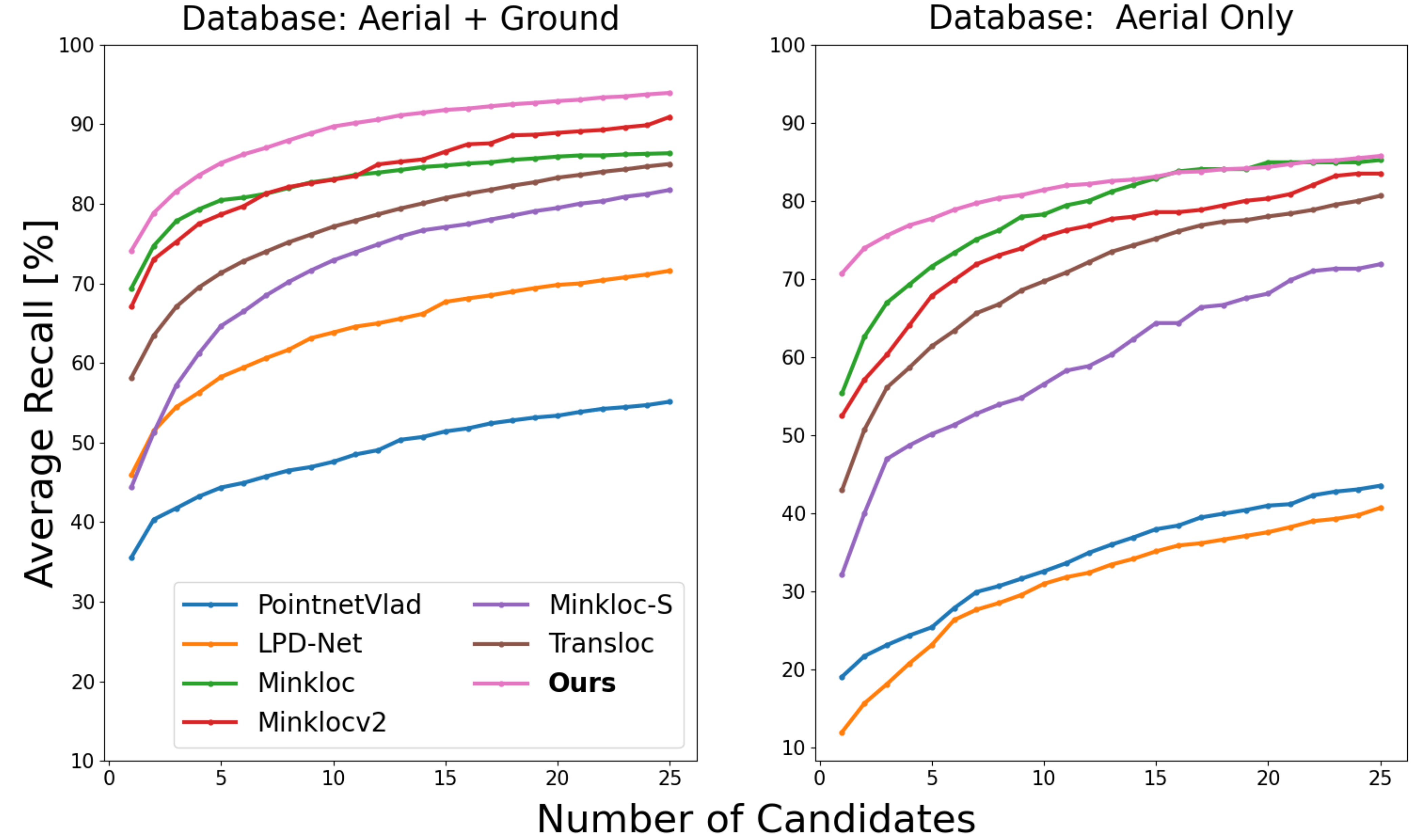}
    \caption{\textbf{\textit{AR@N} curves on \oursdata{}:} Our method performs consistently better than other SOTA methods for top 25 neighbors.}
    \label{fig:ar_curve}
    \vspace{-4mm}
\end{figure}

In Fig.~\ref{fig:ar_curve}, we plot the top 25 predicted neighbors and the corresponding average recall for two evaluation settings in \oursdata{}. There is a clear drop in performance when the database is entirely based on aerial data since the ground-to-aerial match is more difficult. Our method consistently performs better on top 25 candidates retrieval for \textit{aerial + ground} case, and takes lead on top 15 candidates for the \textit{aerial-only} case, compared to other SOTA methods.


\noindent\textbf{Complexity Comparisons:} In Table.~\ref{tab:complexity}, we highlight the model parameter sizes and run-times of different SOTA methods given their performances on \oursdata{} when the database only consists of aerial data.

\begin{table}[t]
\centering
\resizebox{0.95\columnwidth}{!}{%
\begin{tabular}{cccc}
  \toprule
\textbf{Method} & \textbf{AR@1} $\uparrow$  & \textbf{Parameters} $\downarrow$ & \rotatebox[origin=c]{0}{\makecell{\textbf{Running Time} \\\textbf{per Query} $\downarrow$}}
\\

  \midrule

    PointNetVLAD~\cite{pointnetvlad} & 19.07  & 19.78M  & 25 ms\\

    LPD-Net~\cite{lpdnet} & 11.99 & 19.81M & 35 ms  \\
    Minkloc3D~\cite{minkloc3d} & 55.36 & \textbf{1.1M} & \textbf{21 ms} \\    
 \midrule
  
    TransLoc3D~\cite{transloc3d} & 42.97 & 10.97M & 46 ms \\
    TransLoc3D$\dagger$ & 62.57 & 27.85M & 81 ms \\
 \midrule

    \ours{} \textbf{(Ours)} & \textbf{70.73} & 15.35M & 26 ms \\
  \bottomrule
\end{tabular}
}
\caption{\textbf{Performance v.s. Efficiency on \oursdata{}: } Even our model have a relatively large numbers of parameters, we did not compromise much in terms of run-time. $\dagger$ denotes that we manually increase the number of parameters based on the original implementation.}

\label{tab:complexity}

\vspace{-7mm}
\end{table}

\begin{table}[t]
\centering
\resizebox{0.95\columnwidth}{!}{%
\begin{tabular}{ccccc}
  \toprule
\rotatebox[origin=c]{0}{\makecell{\textbf{Num. of Voxel} \\\textbf{Grains ($L$)} }} & \rotatebox[origin=c]{0}{\makecell{\textbf{Quantization} \\\textbf{Size}}} & \rotatebox[origin=c]{0}{\makecell{\textbf{Feature} \\\textbf{Selection}}}   & \textbf{AR@1\%} $\uparrow$ & \textbf{AR@1} $\uparrow$ \\

  \midrule
 \multirow{9}{*}{\Large{1}}  & \multirow{2}{*}{{0.01}}  & \ding{55} & 71.37 & 58.64 \\ 
  &  &    \checkmark & 72.38  & 60.89  \\ 
    
  \cmidrule{2-5}

 & \multirow{2}{*}{{0.08}}  & \ding{55} & 68.93 & 54.10 \\ 
  &  &    \checkmark & 68.77  & 55.57  \\ 
    
  \cmidrule{2-5}
  
 & \multirow{2}{*}{{0.2}}  & \ding{55} & 58.98 & 43.16 \\ 
   & &    \checkmark & 58.32  & 42.11  \\ 
    
  \cmidrule{2-5}

  & \multirow{2}{*}{{0.4}}  & \ding{55} & 54.11 & 37.27 \\ 
  &  &    \checkmark & 51.26  & 35.99  \\ 
    
  \midrule
  
 \multirow{4.5}{*}{\Large{2}}  & \multirow{2}{*}{{0.01 \& 0.08}}  & \ding{55} & 74.76 & 63.85 \\ 
  &  &    \checkmark & 76.22  & 65.75  \\ 

  \cmidrule{2-5}

 & \multirow{2}{*}{{0.01 \& 0.4}}  & \ding{55} & 71.14 & 59.68 \\ 
  &  &    \checkmark & 72.60  & 60.36  \\ 

      \midrule

    \multirow{2}{*}{\Large{3}}  & \multirow{2}{*}{{0.05 \& 0.12 \& 0.4}}  & \ding{55} & 74.69  & 64.33 \\ 
    &   & \checkmark & \textbf{77.47} & \textbf{66.14}  \\ 
    
  \midrule
  
 \multirow{2}{*}{\Large{4}}  & \multirow{1}{*}{\rotatebox[origin=c]{0}{\makecell{0.05 \& 0.12 \\ \& 0.2 \& 0.4}}}  & \ding{55} & 70.97  & 61.07 \\ 
  &  &    \checkmark & 73.22 & 63.14  \\ 
    
  \bottomrule
\end{tabular}
}
\caption{\textbf{Ablation studies on the voxel grains and feature selection: } We achieve the best performance when $L=3$, with $2.78\%$ increase on \textit{AR@1\%} with feature selection.}

\label{tab:grain}

\vspace{-5mm}
\end{table}

\subsection{Ablation Study}
\label{sec:ablation}
\noindent\textbf{Voxel grains and feature selection:} In Table.~\ref{tab:grain}, we show the effect of number of voxel grains $L$ with different quantization size and the feature selection. 
When we only choose one voxel grain, the fine resolution achieves better performance than the coarse one, and the feature selection module does not lead to performance improvement because all information is useful when there are no multi-grained voxelizations to provide potentially repetitive information from different perspectives. As $L$ increases, the effect of feature selection is more significant. The performance start to drop when $L>3$.
One interesting observation is that, as $L$ increases, the model does not require a fine quantization size to achieve good performance, which reduces the computation burden caused by a large $L$.

\begin{table}[t]
\centering
\resizebox{0.93\columnwidth}{!}{%
\begin{tabular}{cccc}
  \toprule
\rotatebox[origin=c]{0}{\makecell{\textbf{Quantization} \\\textbf{Size}}} & \rotatebox[origin=c]{0}{\makecell{\textbf{Refinement} \\\textbf{Process}}}   & \textbf{AR@1\%} $\uparrow$ & \textbf{AR@1} $\uparrow$ \\

  \midrule
 \multirow{3}{*}{\large{0.01 \& 0.08}}  & \ding{55} & 71.03 & 60.43 \\ 
  &    \textit{Fine} $\rightarrow$ \textit{Coarse} & 72.54 & 61.78  \\ 
  &    \textit{Coarse} $\rightarrow$ \textit{Fine} & \textbf{75.28} & \textbf{65.26}  \\ 
    
  \midrule
  
 \multirow{3}{*}{\large{0.05 \& 0.12 \& 0.4}}  & \ding{55} & 70.41 & 60.47 \\ 
  &    \textit{Fine} $\rightarrow$ \textit{Coarse} & 74.96 & 65.56  \\ 
  &    \textit{Coarse} $\rightarrow$ \textit{Fine} & \textbf{77.47} & \textbf{66.14}  \\ 
    
  \midrule
  
 \multirow{3}{*}{\large{ \rotatebox[origin=c]{0}{\makecell{0.05 \& 0.12 \\ \& 0.2 \& 0.4}}}}  & \ding{55} & 71.29 & 59.46 \\ 
  &    \textit{Fine} $\rightarrow$ \textit{Coarse} & 73.22 & 63.14  \\ 
  &    \textit{Coarse} $\rightarrow$ \textit{Fine} & \textbf{77.38} & \textbf{67.15}  \\ 
  
  \bottomrule
\end{tabular}
}
\caption{\textbf{Ablation studies on the refinement process: } We use simple concatenation of multi-grain features after passing through EA blocks~\cite{eanet} individually as a baseline. Refinement starting from coarse to fine leads to $4.25 - 7.06\%$ performance improvement on \textit{AR@1\%}. }

\label{tab:refinement}

\vspace{-6mm}
\end{table}

\noindent\textbf{Refinement process:} In Table.~\ref{tab:refinement}, we evaluate the necessity of the refinement process and different orders of refinement. 
We try starting the refinement process from the finest and coarsest levels and discover that refinement starting from a coarse grain has the most performance improvement, which validates the intuition provided in Sec.~\ref{sec:iterrefine}. 
The coarse representations between different sources are most similar and therefore an easy starting point for metric learning. In the end, the fine-level features can be in a better canonical space and lead to better retrieval results.

\begin{table}[t]
\centering
\resizebox{0.93\columnwidth}{!}{%
\begin{tabular}{cccc}
  \toprule
\rotatebox[origin=c]{0}{\makecell{\textbf{Quantization} \\\textbf{Size}}} & \rotatebox[origin=c]{0}{\makecell{\textbf{ Sub-steps at} \\\textbf{Refinement}}}   & \textbf{AR@1\%} $\uparrow$ & \textbf{AR@1} $\uparrow$ \\

  \midrule
 \multirow{3}{*}{{0.01 \& 0.08}}  & 1 & \textbf{73.99} & 61.81 \\ 
  &    2  & 72.54 & 61.78  \\ 
  &    3 & 72.58 & \textbf{63.41}  \\ 
    
  \midrule
  
 \multirow{3}{*}{{0.05 \& 0.12 \& 0.4}}  & 1 & 73.13 & 63.0 \\ 
  &    2  & \textbf{74.96} & \textbf{65.56}  \\ 
  &    3 & 74.26 & 63.16  \\ 
    
  \midrule
  
 \multirow{3}{*}{{ \rotatebox[origin=c]{0}{\makecell{0.05 \& 0.12 \\ \& 0.2 \& 0.4}}}}  & 1 & 73.70 & 61.51 \\ 
  &    2  & 73.22 & 63.14  \\ 
  &    3 & \textbf{75.61} & \textbf{64.45}  \\ 
  
  \bottomrule
\end{tabular}
}
\caption{\textbf{Ablation studies on the refinement sub-steps: } We see an improvement of $1.45 - 2.39\%$ on \textit{AR@1\%} from adjusting the number of sub-steps.
}

\label{tab:steps}

\vspace{-4mm}
\end{table}

\noindent\textbf{Number of sub-steps during refinement:}
In Table.~\ref{tab:steps}, we run different numbers of sub-steps. Based on the result, we can observe that more refinement sub-steps have better results for a larger number of voxel grains $L$.

\noindent\textbf{Time Embedding:} In Table.~\ref{tab:timeembedding}, we show the effect of time embedding for the refinement. For quantization size $[0.01, 0.08]$, we see a $0.8\%$ decrease  in terms of \textit{AR@1\%}
, which could be explained by the fewer refinement steps $L$ for time embedding to be effective.

\begin{table}[t]
\centering
\resizebox{0.93\columnwidth}{!}{%
\begin{tabular}{cccc}
  \toprule
\rotatebox[origin=c]{0}{\makecell{\textbf{Quantization} \\\textbf{Size}}} & \textbf{Time Embedding}  & \textbf{AR@1\%} $\uparrow$ & \textbf{AR@1} $\uparrow$ \\

  \midrule
 \multirow{2}{*}{{0.01 \& 0.08}}  & \ding{55} & \textbf{73.34} & 61.04 \\ 
  &    \checkmark  & 72.54 & \textbf{61.78}  \\ 
    
  \midrule
  
 \multirow{2}{*}{{0.05 \& 0.12 \& 0.4}}  & \ding{55} & 71.37 & 58.42 \\ 
  &    \checkmark  & \textbf{74.96} & \textbf{65.56}  \\ 
    
  \midrule
  
 \multirow{2}{*}{{ \rotatebox[origin=c]{0}{\makecell{0.05 \& 0.12 \\ \& 0.2 \& 0.4}}}}  & \ding{55} & 72.62 & 59.71 \\ 
  &    \checkmark  & \textbf{73.22} & \textbf{63.14}  \\ 
  
  \bottomrule
\end{tabular}
}
\caption{\textbf{Ablation studies on time embedding: } We notice a trend of improvement with time embedding. 
}

\label{tab:timeembedding}

\vspace{-7mm}
\end{table}

\section{Conclusions, Limitations, and Future Work}

In this paper, we discuss and define the notion of cross-source data, and present a novel cross-source benchmark, \oursdata{}, that poses a new challenge for localization tasks and the vision community in general. 
We present \ours{}, a novel cross-source 3D place recognition method that uses multi-grained features with an iterative refinement process to close the gap between data sources. We show superior performance on the proposed cross-source \oursdata{}, and demonstrate similar performance on the well-established 3D place recognition benchmark Oxford RobotCar, compared to the SOTA methods.


Currently, the task of cross-source 3D place recognition still has room for improvement on the proposed \oursdata{}. In addition, the point registration problem is another challenging topic to pursue. The tasks related to localization and registration using cross-source data open up a series of new challenges based  on our proposed benchmark.


\noindent\textbf{Acknowledgement.} This research was supported by Army Cooperative Agreement No. W911NF2120076 and ARO grant W911NF2110026.

\clearpage

\MakeTitle{
Supplemental Material \\
\textsc{CrossLoc3D}:
Aerial-Ground Cross-Source 3D Place Recognition 
}{}{}

\section{More Details on \oursdata{} Collection}

\begin{figure}[b]
    \centering
    \includegraphics[width=\columnwidth]{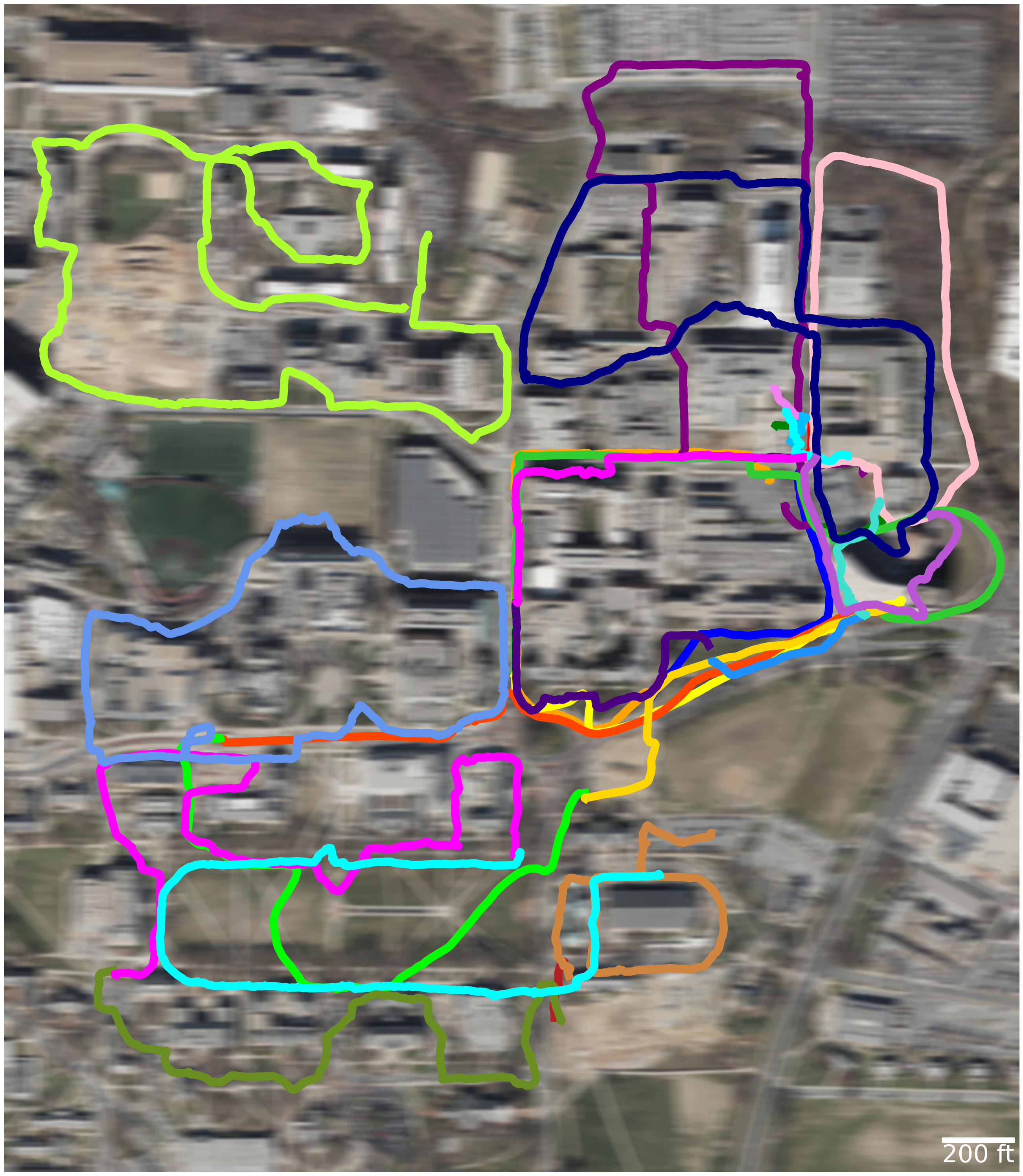}
    \caption{\textbf{\oursdata{} Routes.}}
    \label{fig:coverage}
\end{figure}

In Fig.~\ref{fig:coverage}, we show the routes for the ground LiDAR scan collection. We blur the satellite imagery of the collection location due to the anonymity policy. The total length of the collection is roughly 247 minutes, and we remove the LiDAR scan in the location where the covariance of the GPS location exceeds $10\ m$ due to poor satellite signal. 

The aerial data covers the entire region, while the ground data is sparsely distributed along the routes. 
One of our main goals is to find the correspondence between the features of the aerial (dense) and ground (sparse) datasets.
During training, we include the aerial data as a database, which corresponds to the region in the training split of the ground data. During testing, all query points are taken from a disjoint set of ground data.
In contrast to \cite{oxford}, we include all aerial data in the testing database, instead of a disjoint subset of the aerial data. (In practice, if we narrow down the query location to a small subset of the database like \cite{oxford}, it defeats the purpose of our data retrieval task.)

\section{Implementation Details on \ours{}}

In this section, we add more details about the implementation of \ours{}.

\noindent\textbf{Backbone: } For each voxel set $V_i$, we first perform a sparse convolution operation with a kernel size of 5 and a stride of 1, and another sparse convolution operation with a kernel size of 3 and a stride of 2, before multi-scale sparse convolution. The output feature size of both sparse convolutions is 64, and each sparse convolution is followed by a batch normalization, and a ReLU activation.

\begin{figure*}[t]
    \centering
    \includegraphics[width=\textwidth]{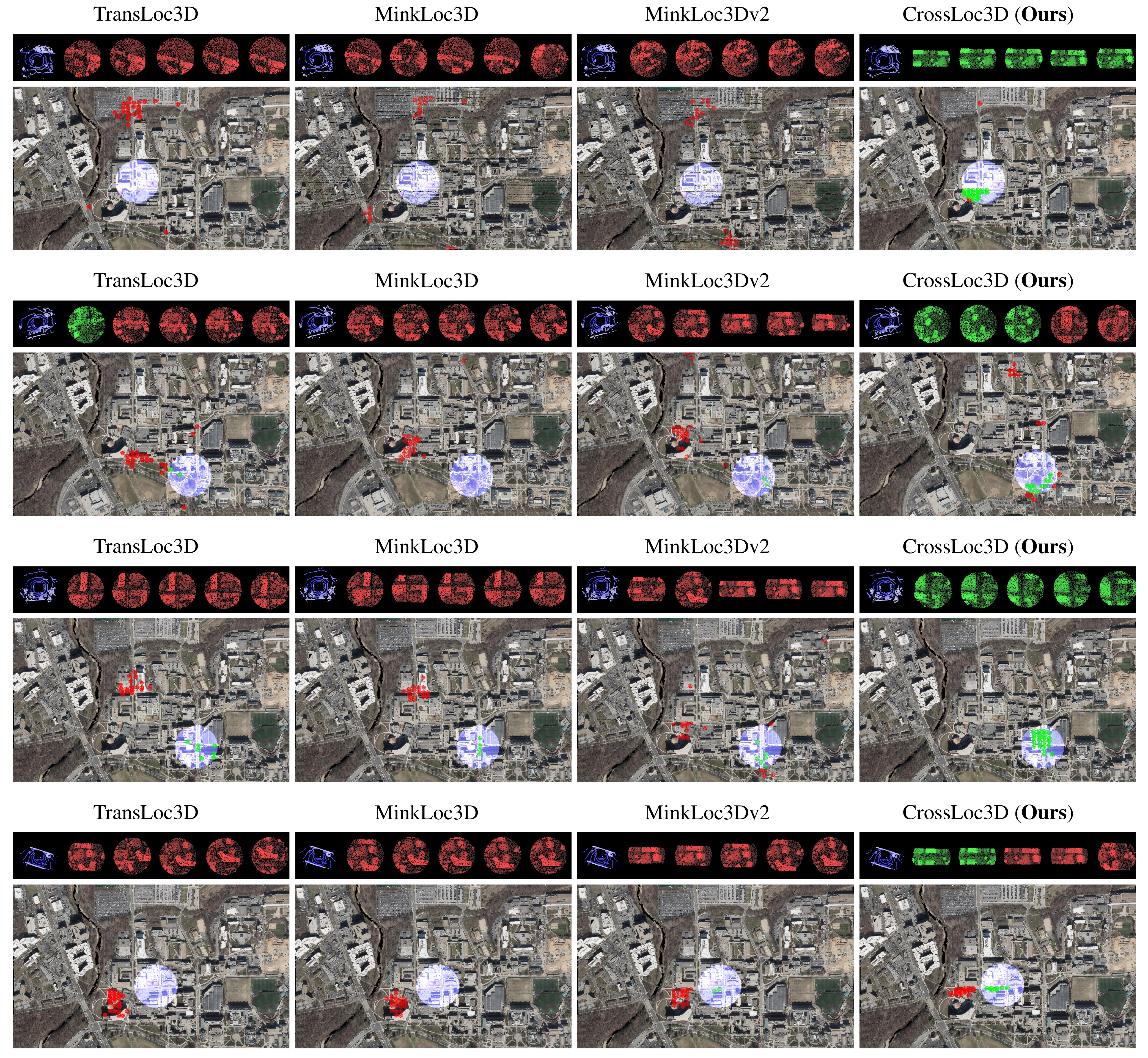}
    \caption{\textbf{More qualitative comparisons between \ours{} and other SOTA methods~\cite{transloc3d, minkloc3d, minkloc3dv2}:} \textbf{For each row, top:} Input ground query (\textcolor{blue}{blue}) and top 5 retrievals (ranked from left to right) from the database of aerial data (true neighbor -- \textcolor{green}{green}, false neighbor -- \textcolor{red}{red}). \textbf{For each row, bottom:} Distributions of ground query and top 25 retrievals. 
    }
    \label{fig:inf_vis}
    \vspace{-7pt}
\end{figure*}

\noindent\textbf{Iterative Refinement and NetVLAD: } The EA block has a attention map that increases in linear complexity compared to the traditional quadratic complexity in transformer architecture. A single stream of the feature $\hat{F}$, which is also the query in traditional transformer, is passed as input. The self query is correlated with a pre-trained memory unit, which after training, contains the context of the entire point cloud dataset. The attention map's normalization allows scale invariant feature vectors. For EA blocks~\cite{eanet}, we use 4 attention heads and a feature size of 64. We use two linear layers and one GELU activation to generate the time embedding. Following the iterative refinement, we use a 1D convolution with a feature size of 512 and a kernel size of 1, a batch normalization and a ReLU activation. For NetVLAD, the number of clusters $K$ is 64.


\noindent\textbf{\ours{} for Oxford RobotCar: }We use a different configuration setting for the single-source benchmark, Oxford RobotCar~\cite{oxford}, compared to the configuration for the proposed \oursdata{}. We use a quantization size of $[0.01, 0.12, 0.2]$ with a sub-step size of 2. In addition, the feature dimension of the NetVLAD is 512.

\begin{figure*}[t]
    \centering
    \includegraphics[width=\textwidth]{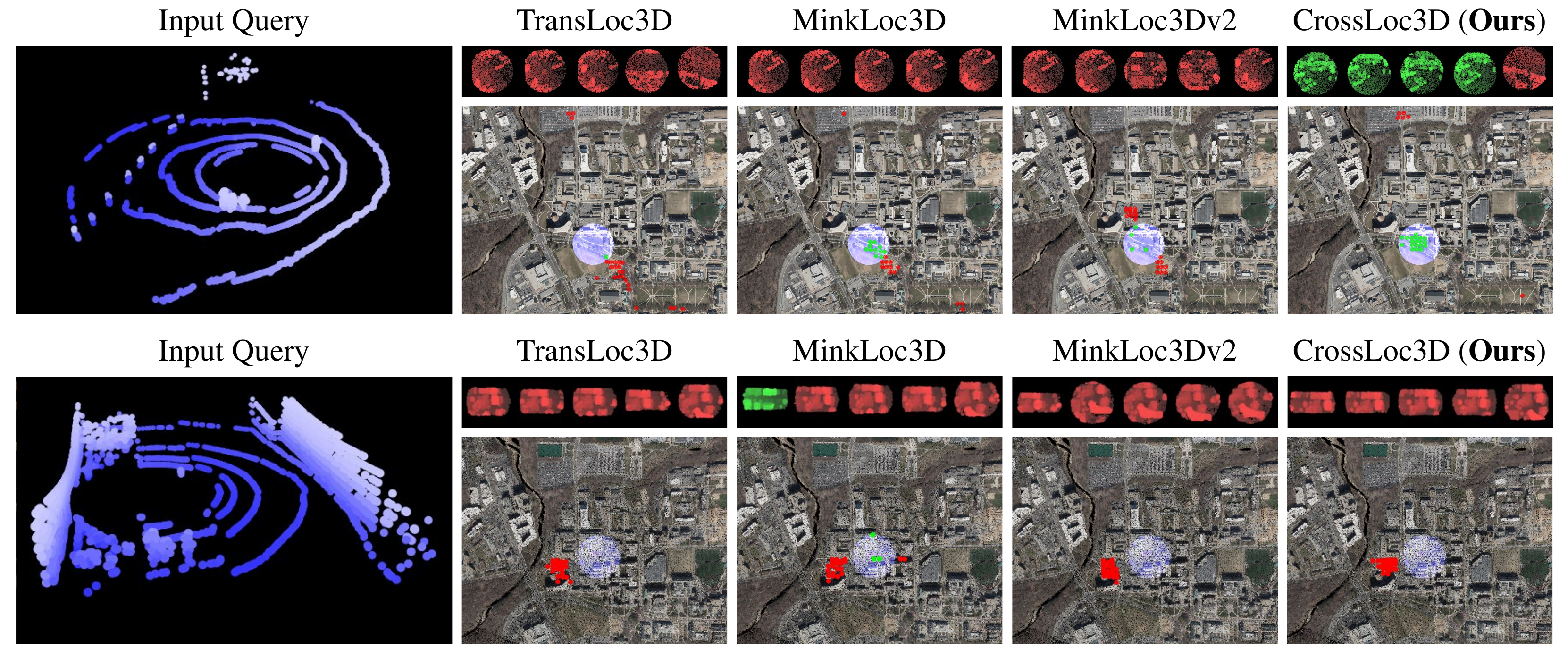}
    \caption{\textbf{Two Challenging cases for \ours{} and other SOTA methods~\cite{transloc3d, minkloc3d, minkloc3dv2}:} \textbf{For each row, left:} Input ground query (\textcolor{blue}{blue}). \textbf{For each row, top:} Top 5 retrievals (ranked from left to right) from the database of aerial data (true neighbor -- \textcolor{green}{green}, false neighbor -- \textcolor{red}{red}). \textbf{For each row, bottom:} Distributions of ground query and top 25 retrievals. 
    }
    \label{fig:failure_case}
\end{figure*}

\section{Inference Visualizations}
\label{sec:inf_vis}

We give some visualization results on \oursdata{} in Fig.~\ref{fig:inf_vis}. In Fig.~\ref{fig:failure_case}, we show two challenging queries in the \oursdata{} benchmark. In the example on the top, the scan is captured in an open area with limited ground features, which causes some issues for retrieval. In the second example, the top 25 recall is low due to the lack of distinctive features on the ground, except for two parallel walls.

{\small
\bibliographystyle{ieee_fullname}
\bibliography{egbib}
}

\end{document}